\documentclass{article}

\usepackage{microtype}
\usepackage{graphicx}
\usepackage{subfigure}
\usepackage{booktabs} % for professional tables
\usepackage{amsmath,amssymb,amsfonts,amsthm}

\newtheorem{theorem}{Theorem}
\newtheorem{assumption}{Assumption}

\usepackage{hyperref}

\usepackage[accepted]{icml2021}

\icmltitlerunning{Learned Interpretable Residual Extragradient ISTA for Sparse
Coding}

\begin{document}

\twocolumn[
\icmltitle{Learned Interpretable Residual Extragradient ISTA for Sparse Coding}



\icmlsetsymbol{equal}{*}

\begin{icmlauthorlist}
\icmlauthor{Lin Kong}{mig}
\icmlauthor{Wei Sun}{mig}
\icmlauthor{Fanhua Shang}{mig,pc}
\icmlauthor{Yuanyuan Liu}{mig}
\icmlauthor{Hongying Liu}{mig}
\end{icmlauthorlist}

\icmlaffiliation{mig}{Key Lab.\ of Intelligent Perception and Image
	Understanding of Ministry of Education, School of Artificial Intelligence,
	Xidian University, China.}
\icmlaffiliation{pc}{Peng Cheng Laboratory, Shenzhen, China}

\icmlcorrespondingauthor{Fanhua Shang}{fhshang@xidian.edu.cn}

\icmlkeywords{Algorithm Unrolling, Sparse Coding, Extragradient, ResNet, Optimization}

\vskip 0.3in
]



\printAffiliationsAndNotice{}  % leave blank if no need to mention equal
%contribution
%\printAffiliationsAndNotice{\icmlEqualContribution} % otherwise use the
%%standard text.

\begin{abstract}
   Recently, the study on learned iterative shrinkage thresholding algorithm
   (LISTA) has attracted increasing attentions. A large number of experiments as
   well as some theories have proved the high efficiency of LISTA for solving
   sparse coding problems. However, existing LISTA methods are all serial
   connection. To address this issue, we propose a novel extragradient based
   LISTA
   (ELISTA), which has a  residual structure and theoretical guarantees. In  particular, our algorithm can also provide the interpretability for Res-Net to
   a certain extent.
   From a theoretical perspective, we prove
   that our method attains linear convergence. In practice, extensive empirical results verify the advantages of our method.
\end{abstract}

\section{Introduction}
\label{intro}
In this paper, we mainly consider the following problem, which is to
recover a sparse vector $x^*\in\mathbb{R}^n$ from an observation vector
$y\in\mathbb{R}^m$ with noise $\varepsilon\in\mathbb{R}^m$ (e.g., additive
Gaussian white noise):
\begin{equation}\label{eq1}
y=Ax^*+\varepsilon,
\end{equation}
where $A\in\mathbb{R}^{m\times n}$ ($m\ll n$ in general) is the dictionary
matrix. To solve Problem \eqref{eq1} which is generally ill-posed, some prior
information such as sparsity or low-rankness needs to be incorporated, for
example, $x^*$ is sparse. A
common
way to estimate $x^*$ is to solve the Lasso problem \cite{lasso1996}:
\begin{equation}\label{lasso}
\mathop{\min}\limits_{x\in\mathbb{R}^n}P(x)=f(x)+g(x)=\frac{1}{2}\|y-Ax\|_2^2+\lambda\|x\|_1,
\end{equation}
where $\lambda\geq0$ is a regularization parameter. Many methods have been
proposed to solve the sparse coding problem, such as least angle regression
\cite{lar2004}, approximate message passing (AMP)
\cite{amp2009} and iterative shrinkage thresholding algorithm (ISTA)
\cite{ista2004, ista2008}. For solving Problem \eqref{lasso}, the update rule of ISTA is
	\begin{equation*}\label{ista}
x^{t+1}=\mathrm{ST}\Big(x^t+\frac{1}{L}A^T(y-Ax^t),\frac{\lambda}{L}\Big),\;\;
t=0,1,2,...,
\end{equation*}
where $\mathrm{ST}(\cdot,\theta)$ is the soft-thresholding (ST) operator
with
the threshold $\theta$, $\frac{1}{L}$ is the step size which should be
taken in
$(0, \frac{2}{L})$, where  $L$ is the largest singular value of the
dictionary 	matrix. \citet{ista2009} proved that ISTA can only achieve
a sublinear convergence rate.

Recently, a class of methods of unfolding the traditional iterative algorithms
into deep neural networks (DNNs), which are called \textit{Algorithm Unfolding} \cite{Monga2021unrolling} or
\textit{Deep Unfolding}  \cite{Hershey2014},
have been proposed, and have gradually attracted more and more attention.
This idea was first proposed by \citet{lista}, and they unfolded ISTA and
viewed
ISTA as a recurrent neural network (RNN) and proposed a
learning-based model named Learned ISTA (LISTA):
\begin{equation}\label{lista}
\begin{split}
x^{t+1}=\mathrm{ST}(W_1^ty+W_2^tx^t,\theta^t),\quad t=0,1,2,...,
\end{split}
\end{equation}
where $W_1^t$, $W_2^t$ and $\theta^t$ are initialized as $\frac{1}{L}A^T$,
$I -
\frac{1}{L}A^TA$ and $\frac{\lambda}{L}$, respectively. All the parameters
$\Theta = \{W_1^t, W_2^t, \theta^t\}$ are learnable and data-driven. Many
empirical and theoretical results as in \cite{2020Ada, giryes2018tradeoffs} have
shown that  LISTA can recover $x^*$ from $y$ more
accurately and use one or two order-of-magnitude fewer iterations than
original ISTA. Moreover, the linear convergence of a variant of LISTA
(i.e., LISTA-CPSS) was proved for the first time in \cite{lista_cpss}. In
addition, these networks have higher interpretability than
general networks, thus can provide some explanations for
deep networks. Actually, the deep unfolding algorithm (actually a network)
was believed to incorporate
some priors of models and algorithms in traditional optimization problems and
have the learning capacity of network obtained from training data.

Due to the advantages of the idea of algorithm unfolding, a lot of works such
as
\cite{wang2016l0,
	learnlow2015, TISTA, lamp, lcsc2018}
inspired by \cite{lista} have been proposed and successfully applied in
various fields. Moreover, a series of studies on LISTA have attracted
increasing attentions and
inspired
many subsequent works  in different aspects, including learning based
optimization \cite{dladmm2019, admmnet2016}, design of DNNs \cite{
	ldamp2017,
	istanet2018, sc2net2018, rna2020, rick2017one,
	zhang2020novel,simon2019rethinking} and interpreting the DNNs
\cite{zarka2019deep, papyan2017convolutional, aberdam2019multi,
	sulam2018multilayer, sulam2019multi}.

 There are also many works such as \cite{xin2016maximal,
giryes2018tradeoffs,
	underlista2016, lista_cpss, alista, glista, ablin2019learning} to discuss
	and understand
LISTA and
its variants from a theoretical perspective. Among them, \citet{lista_cpss}
proved that there is a coupling relationship between the two learnable matrices
of each layer of LISTA, thereby reducing the number of learnable parameters.
They also proved the linear convergence of LISTA for the first time.
Later, many subsequent works \cite{alista, glista, ablin2019learning} further
improved LISTA with different methods. For instance, \citet{alista} simplified
the different matrix parameters of each layer of the network to the product of
a matrix shared by the network and different scalar parameters of each layer,
and proved that using the matrix parameters obtained by solving an
optimization problem can achieve the same performance obtaind by learnable
matrices. Then \citet{glista} proposed that the value of the element in the
estimate obtained by LISTA may be lower than the expected value, and thus,
inspired by gated
recurrent unit (GRU) \cite{cho2014learning, chung2015gated}, GLISTA
\cite{glista} was proposed
to gain the LISTA-related algorithms. Besides, we also make improvements based
on LISTA and proposed an innovative work \cite{elista}, and this paper is a
condensed version of \cite{elista}.

However, we find that all the existing variants of LISTA with convergence
guarantees are
serial, the residual network (Res-Net) \cite{he2016deep}, which is
influential
in deep learning, has not been introduced into LISTA. An important reason is
that changing the original structure of LISTA will destroy its excellent
mathematical interpretability. Can we get a new LISTA with
an interpretable residual structure, which has a convergence guarantee?

\textbf{Our Main Contributions:} The main contributions of this paper are
listed	as follows:

$\bullet$ We propose a novel unfolding network, named Extragradient based LISTA (ELISTA), which is a variant of  LISTA with  residual structure by employing the idea of extragradient into LISTA and establishing the relationship with Res-Net,
which is an improvment about the network structure  for solving sparse coding
problems. To the best of our knowledge, this is  the first residual
structure LISTA with theoretical guarantees.

	$\bullet$ We prove the linear convergence of ELISTA. Moreover, we conduct extensive experiments to verify the
effectiveness of our algorithm. The experimental results
show that our ELISTA is superior to the state-of-the-art methods.

\section{Extragradient Based LISTA}\label{alg}
In this section, we first introduce the technique of extragradient into
LISTA and propose an innovative algorithm, named \textit{Extragradient based
LISTA}
(ELISTA), and depict it in detail. Moreover, we establish the
relationship between ELISTA and Res-Net, which is one of the reasons why
ELISTA is advantageous.

\subsection{Extragradient Method}\label{extrag}
We note that iterative algorithms, such as ISTA, can actually be treated as
a proximal gradient descent method, which is a first-order optimization
algorithm, for special objective functions. Thus, we want to introduce the
idea of extragradient into the related iterative algorithms. The
extragradient method was first proposed by
\cite{korpelevich1976extragradient}, which is a classical method for
variational inequality problems. For optimization problems, the idea of
extragradient was first used in \cite{nguyen2018extragradient}, which
proposed an extended extragradient method (EEG) by combining this idea with
some first-order descent methods. In the $t$-th iteration of EEG, it first
calculates the gradient at $x^t$, and updates $x^t$ according to the
gradient to get an intermediate point $x^{t+\frac{1}{2}}$, then calculates the
gradient at $x^{t+\frac{1}{2}}$, and updates the original point $x^t$
according to the gradient at the intermediate point $x^{t+\frac{1}{2}}$ to
obtain
$x^{t+1}$, which  is the key idea of extragradient. Intuitively, the
additional step in each iteration of EEG allows us to examine the geometry
of the problem and consider its curvature information, which is one of the
most important bottlenecks for first-order methods. Thus, by using the idea
of extragradient, we can get a better result after each iteration. The
update rules of EEG for Problem \eqref{lasso} can be rewritten as follows:
\begin{equation}\label{update}
\begin{array}{c}
x^{t+\frac{1}{2}} = \mathrm{ST}\left(x^{t} - \frac{1}{L}A^T(Ax^t -
y),\frac{\lambda}{L}\right), \\
\;x^{t+1} = \mathrm{ST}\Big(x^{t} - \frac{1}{L}A^T(Ax^{t+\frac{1}{2}} -
y),\frac{\lambda}{L}\Big).
\end{array}
\end{equation}
This form of EEG is similar to ISTA, and thus it can be regarded as a
generalization of ISTA.

\subsection{Extragradient Based LISTA and the Relationship with Res-Net}

\begin{figure*}[t]
	\centering
	\subfigure[Res-Net: a building
	block.]{\includegraphics[width=0.5\columnwidth]{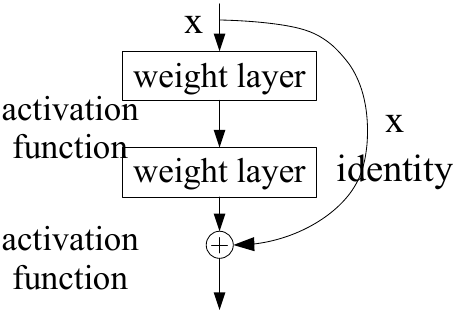}} \;\;\;\;\quad
	\subfigure[ELISTA: a building
	layer.]{\includegraphics[width=1.3\columnwidth]{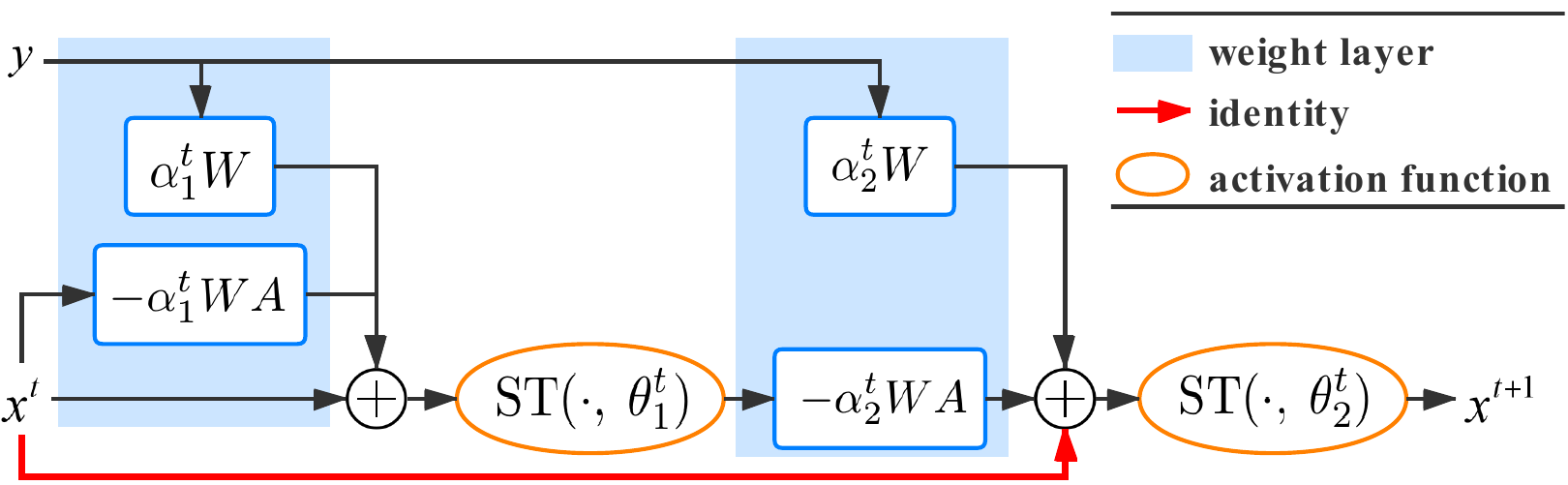}}
	\vspace{-1mm}
	\caption{Comparison of the network structures of Res-Net \cite{he2016deep} and ELISTA (ours).}
	\label{evsr}
	\vspace{-1mm}
\end{figure*}

In order to speed up the convergence of EEG, we combine the algorithm with
deep networks and regard $\frac{1}{L}A^T$ and two thresholds of two steps
in \eqref{update} as learnable parameters, and get the following update
rules:
\begin{equation}\label{ureeg}
\begin{array}{c}
x^{t+\frac{1}{2}} = \mathrm{ST}(x^t-W_1^t(Ax^t-y),\theta_1^t), \\
x^{t+1} = \mathrm{ST}(x^t-W_2^t(Ax^{t+\frac{1}{2}}-y),\theta_2^t).
\end{array}
\end{equation}
However, since the above scheme has two different matrices $W_1^t$ and
$W_2^t$
to learn in each layer, the number of network parameters greatly increases
and
the training of the network slows down significantly. Therefore, to address
this issue and further establish the connection between the two steps of
\eqref{ureeg}, we convert $W_1^t$ and $W_2^t$ into $\alpha_1^tW^t$ and
$\alpha_2^tW^t$, respectively, where $\alpha_1^t$ and $\alpha_2^t$ are two
scalars to learn. Then, inspired by \cite{alista}, we change the $W^t$ of
each
layer into the same $W$ and get a tied algorithm, which can significantly
reduce the number of learnable parameters.
Finally, we
obtain the following update rules for our \textit{Extragradient Based LISTA}
(ELISTA):
\begin{equation}\label{ur}
\begin{array}{c}
x^{t+\frac{1}{2}} = \mathrm{ST}(x^t-\alpha_1^tW(Ax^t-y),\ \theta_1^t), \\
x^{t+1} = \mathrm{ST}(x^t-\alpha_2^tW(Ax^{t+\frac{1}{2}}-y),\ \theta_2^t).
\end{array}
\end{equation}

According to \eqref{ur}, we can get the network structure diagram
of
ELISTA, as shown in Figure \ref{evsr}. Through our observation and comparison,
we find that the network
structure of ELISTA is corresponding to Res-Net. Since $y$ is already
given, we can regard $y$ as a bias. Thus, from Figure \ref{evsr}, we can see
that the structure of the network obtained by ELISTA is the same as that of
Res-Net, including weight layer, activation function and identity. As we all
know, Res-Net can obtain a better performance by improving network
structures. Therefore, it is meaningful to discuss and study the explanation
for
the internal mathematical mechanism of Res-Net. On the one hand, to some
extend, our algorithm may be regarded as a mathematical explanation of the
reason for the superiority of Res-Net. On the other hand, the connection and
combination of ELISTA and Res-Net might be able to explain why our algorithm
has better performance than existing methods. Besides, there are a lot of
work using ordinary differential equation (ODE) to interpret the network by
considering ODE as a continuous equivalent of the residual network (ResNet)
\cite{chen2018neural}. However, we found that ODE can only explain the networks
with linear connection blocks, while ours is nonlinear. But, the form of our
blocks are less general than those of ODE.

\begin{table}[t]
	\small
	\centering
	\setlength{\tabcolsep}{2.5pt}
	\renewcommand\arraystretch{1.3}
	\caption{\fontsize{10pt}{0} Comparison of the number of parameters to learn
		in different methods.} 	\label{table_num_of_par}
	\begin{tabular}{cccc}
		\toprule[2pt]
		
		LISTA     & LAMP  & GLISTA  & ELISTA \\
		\hline
		$\mathcal{O}(TMN\!+\!T)$ & $\mathcal{O}(TMN\!+\!T)$ &
		$\mathcal{O}(TMN\!+\!T)$ & $\mathcal{O}(MN\!+\!T)$\\
		\bottomrule[2pt]
	\end{tabular}
\vspace{-2mm}
\end{table}

Moreover, the comparison of the number of parameters of the network
corresponding to different algorithms is shown in Table
\ref{table_num_of_par}, where LAMP \cite{lamp} is an algorithm to transform
AMP \cite{amp2009} into a neural network inspired by \cite{lista}.

\section{Convergence Analysis}
In this section, we provide the convergence analysis of our algorithm. We
first give a basic assumption. Then we provide the
convergence property of ELISTA. We note that
our analysis, like that of Theorems 3 and 4 of \cite{glista}, is proved
under the existence of ``false positive", while the theoretical analysis of
\cite{lista_cpss, alista} was provided under the assumption of no ``false
positive", which is difficult to satisfy in reality.

	\begin{assumption}[Basic assumption]\label{ass1}
	The signal $x^*$ is sampled from the following set:
	\begin{equation*}
	x^*\in\mathcal{X}(B,s)\,\mathop{\rm{=}}\limits^{\triangle}\,\{x^*||x_i^*|\leq
	B,\forall i,\|x^*\|_0\leq s\}.
	\end{equation*}
	In other words, $x^*$ is bounded and $s$-sparse $(s\geq 2)$.
	Furthermore, we assume $\varepsilon=0$.
\end{assumption}

This assumption is a basic assumption for this class of algorithms. Almost all the related algorithms need to satisfy this
assumption, e.g., \cite{alista,glista}.

	Based on the assumption, we can get the linear
convergence of ELISTA, which can be given by the following theorem.

\begin{theorem}[Linear Convergence for ELISTA]\label{thl}
	If Assumption \ref{ass1} holds, $W^t\in \mathcal{W}(A)$ can be satisfied
	by selecting $W^t$ properly,
	\begin{equation}\label{thetaut}
	\begin{array}{c}
	\theta_1^t =
	\alpha_1^t\omega_{t+\frac{1}{2}}(k_{t+\frac{1}{2}}|\Theta)\mu(A)\sup_{x^*}\|x^t
	- x^*\|_1,
	\\
	\theta_2^t =
	\alpha_2^t\omega_{t+1}(k_{t+1}|\Theta)\mu(A)\sup_{x^*}\|x^{t+\frac{1}{2}}
	-
	x^*\|_1
	\end{array}
	\end{equation}
	are achieved, $\alpha_1^t,\alpha_2^t\in(0,\frac{2}{1+(2s-1)\mu(A)})$
	and $s$ is small enough, then for sequences generated by ELISTA, there
	exist ``false positive" with $0<k_{t},k_{t+\frac{1}{2}}<s$ and
	\begin{equation}\label{thlinear}
	\|x^t - x^*\|_2\leq sB\exp\Big(\sum_{i=1}^{t}c_i^*\Big)<sB\exp(ct),
	\end{equation}
	where $c_i^*<0$, and $c = \max_{i=1,2,...,t}\{c_i^*\}<0$.
\end{theorem}
The definitions of $\mathcal{W}(A)$ and $\mu(A)$ can be found in Definition 1
in
\cite{alista}. From
Lemma 1 in \cite{lista_cpss}, we know $\mathcal{W}(A)\neq\varnothing$. Besides,
the definitions of $\omega_{t+\frac{1}{2}}(k_{t+\frac{1}{2}}|\Theta)$ and
$\omega_{t+1}(k_{t+1}|\Theta)$ can be given by referring to Definition 2 in
\cite{glista}. Theorem \ref{thl} shows that our ELISTA attains
linear convergence. We note that we have not given the detailed proof of
Theorem \ref{thl}, due to page limits. We will provide it in our future
work.

\section{Experimental Results}
In this section, we evaluate our ELISTA in terms of sparse representation
performance and 3D geometry recovery via photometric stereo. All
the experimental settings are
the same as the previous works \cite{lista_cpss,alista,glista}. However, the
performance of SS \cite{lista_cpss} is
greatly
affected by the hyper-parameters, and it is necessary to know the sparsity
of
$x^*$ in advance to set the hyper-parameters, which is difficult to get in
real
situations. Thus, in order to more fairly compare the impact of the network
itself on performance, all the networks do not use SS. All training follows
\cite{lista_cpss}.
For
all the methods, $\alpha_1^t$ and $\alpha_2^t$ are initialized as 1.0, and
$\theta_1^t$
and
$\theta_2^t$ are initialized as $\frac{\lambda}{L}$. All the results are
obtained by running
ten
times and averaged.

\subsection{Sparse Representation Performance} \label{exp_sub1}
In this subsection, we compare our ELISTA with the
state-of-the-art methods: LISTA, LAMP
and GLISTA.  We set $m \!= \!250$, $n \!=\! 500$ and
$T \!=\!16$, and train the networks with two different noise
levels: SNR (Signal-to-Noise Ratio) = 30, $\infty$ and three different ill
conditioned matrices $A$ with condition numbers $\kappa$ = 5, 50, 500. For
detailed data generation methods, please see
\cite{elista}.

\begin{table}
	\centering
	\setlength{\tabcolsep}{2pt}
	\renewcommand\arraystretch{1.0}
	\caption{Comparison of the NMSE performance with different algorithms
		under different $\kappa$ and SNR.} \label{table_snr_col}
	\begin{tabular}{c|cccc}
		\toprule[2pt]
		       & LISTA     & LAMP   &
		      GLISTA & ELISTA \\
		\hline
	$\kappa = 5$, SNR $=\infty$ & -38.658 & -44.967 & -65.569 &
	\textbf{-83.997}\\
			$\kappa = 50$, SNR $=\infty$ & -37.471 & -46.385 &
			-63.523 &
		\textbf{-82.848}\\
		$\kappa=500$, SNR $=\infty$  &-31.845 & -43.097 & -57.542 &
		\textbf{-77.865}\\
		$\kappa=500$, SNR $=30$ & -23.593  & -25.045 & -32.757 &
		\textbf{-32.832}\\
		\bottomrule[2pt]
			\end{tabular}
	\vspace{-3mm}
\end{table}

Table \ref{table_snr_col} shows that our method obviously outperform
the
compared methods in the
noiseless case. Especially, compared with LISTA,  the NMSE performance of
our
method is almost twice as much as that of LISTA. In the presence of noise,
our method achieves the state-of-the-art accuracy.

\vspace{-2mm}
\subsection{3D Geometry Recovery via Photometric Stereo}
\begin{table}
	\centering
	\setlength{\tabcolsep}{15.3pt}
	\caption{\fontsize{10pt}{0} The mean angular error of 3D geometry recovery
		via photometric stereo.} \label{table_3d}
	\begin{tabular}{ccccc}
		\toprule[2pt]
		$q$     & LISTA     & GLISTA   & ELISTA \\
		\hline
		35 & 0.06836 & 0.06249 & \textbf{0.04724}\\
		25 & 0.09664 & 0.10033 & \textbf{0.06597}\\
		15 & 0.69334 & 0.63967 & \textbf{0.53269}\\
		\bottomrule[2pt]
		
	\end{tabular}
	\vspace{-8mm}
\end{table}

In this subsection, we compare our ELISTA with the
state-of-the-art methods: LISTA and GLISTA for 3D geometry recovery via
photometric stereo, which is a powerful technique used to recover high
resolution
surface
normals from a 3D scene using appearance changes of 2D images in different
lighting \cite{woodham1980photometric}. In practice, however, the
estimation
process is often interrupted by non-lambert effects, such as highlights,
shadows, or image noise. This problem can be solved by decomposing the
observation matrix of the superimposed image under different lighting
conditions into ideal lambert components and sparse error terms
\cite{wu2010robust, ikehata2012robust}, i.e., $o=\rho Lw+e$, where $o \in
\mathbb{R}^q$ denotes the resulting measurements, $w \in \mathbb{R}^3$
denotes
the true surface normal, $L \in \mathbb{R}^{q\times 3}$ defines a lighting
direction, $\rho$ is the diffuse albedo, acting here as a scalar multiplier
and
$e \in \mathbb{R}^q$ is an unknown sparse vector. By multiplying both sides
of
$o=\rho Lw+e$ by the orthogonal complement to $L$, we can get
$Proj_{null_{[L^\top]}}(o) = Proj_{null_{[L^\top]}}(e)$. Let
$Proj_{null_{[L^\top]}}(o)$ be $y$ and $Proj_{null_{[L^\top]}}(e)$ be $Ax$,
$e$
can be obtained by solving the sparse coding problem. Then we can use
$L^\dagger(o-e)$ to recover $n$. The main experimental settings follow
\cite{xin2016maximal, glista, he2017bayesian}. Tests are performed using the
32-bit HDR gray-scale images of objects ``Bunny" as in
\cite{xin2016maximal} with $q = 35, 25, 15$ and 40$\%$ of the
elements of the
sparse noise $e$ are non-zero. From Table \ref{table_3d}, we can find that
our
method performs much better than LISTA and GLISTA.

\vspace{-3mm}

	\section{Conclusions}
%we consider a sparse representation problem.
We  proposed a novel extragradient based learned iterative
shrinkage thresholding algorithm (called ELISTA) with an interpretable residual
structure. Moreover, we proved ELISTA can achieve linear convergence.
Extensive empirical results verified the high efficiency of our method. This
could have both theoretical and practical impacts to the relationship between
new neural network architectures and advanced algorithms, and potentially
deepen our understanding to interpretability of deep learning models. One
limitation of this paper is that we use the same assumption as in
the previous work \cite{lista_cpss, alista, glista}, that the sparsity  of
$x^*$ is small enough. Removing this common assumption is our future work.

\section*{Acknowledgments}
This work was supported by the National Natural Science Foundation of China (Nos.\ 61876221, 61876220 and 61976164), the Project supported the Foundation for Innovative Research Groups of the National Natural Science Foundation of China (No.\ 61621005), the Major Research Plan of the National Natural Science Foundation of China (Nos.\ 91438201 and 91438103), the Program for Cheung Kong Scholars and Innovative Research Team in University (No.\ IRT\_15R53), the Fund for Foreign Scholars in University Research and Teaching Programs (the 111 Project) (No.\ B07048), and the National Science Basic Research Plan in Shaanxi Province of China (No.\ 2020JM-194).

% Note use of \abovespace and \belowspace to get reasonable spacing
% above and below tabular lines.

%\section*{Acknowledgements}

\nocite{langley00}

\bibliography{ELISTA_ICMLworkshop}
\bibliographystyle{icml2021}

\end{document}